\def\year{2020}\relax
\documentclass[letterpaper]{article} % DO NOT CHANGE THIS
\usepackage{aaai20}  % DO NOT CHANGE THIS
\usepackage{times}  % DO NOT CHANGE THIS
\usepackage{helvet} % DO NOT CHANGE THIS
\usepackage{courier}  % DO NOT CHANGE THIS
\usepackage[hyphens]{url}  % DO NOT CHANGE THIS
\usepackage{graphicx} % DO NOT CHANGE THIS
\usepackage{graphicx}  % DO NOT CHANGE THIS
\usepackage[section]{placeins}
\usepackage{amsmath}
\usepackage{amssymb}
\usepackage{subcaption}
\title{Discrete and Continuous Action Representation for Practical RL in Video Games}
\author{Olivier Delalleau\textsuperscript{*}\textsuperscript{\rm 1}, Maxim Peter\textsuperscript{*}, Eloi Alonso, Adrien Logut\\ % All authors must be in the same font size and format. Use \Large and \textbf to achieve this result when breaking a line
Ubisoft La Forge %If you have multiple authors and multiple affiliations
}

\renewcommand{\H}{\mathcal{H}}
\newcommand{\R}{\mathbb{R}}
\newcommand{\citet}{\cite}  % for now use \cite everywhere
\newcommand{\citep}{\cite} % for now use \cite everywhere
\DeclareMathOperator*{\softmax}{softmax}

\begin{document}

\maketitle

\begin{abstract}
While most current research in Reinforcement Learning (RL) focuses on improving the performance of the algorithms in controlled environments, the use of RL under constraints like those met in the video game industry is rarely studied. Operating under such constraints, we propose Hybrid SAC, an extension of the Soft Actor-Critic algorithm able to handle discrete, continuous and parameterized actions in a principled way. We show that Hybrid SAC can successfully solve a high-speed driving task in one of our games, and is competitive with the state-of-the-art on parameterized actions benchmark tasks. We also explore the impact of using normalizing flows to enrich the expressiveness of the policy at minimal computational cost, and identify a potential undesired effect of SAC when used with normalizing flows, that may be addressed by optimizing a different objective.
\end{abstract}

\section{Introduction}

Reinforcement Learning (RL) applications in video games have recently seen massive advances coming from the research community, with agents trained to play Atari games from pixels \cite{mnih2015human} or to be competitive with the best players in the world in complicated imperfect information games like DOTA 2 \cite{openaifive} or StarCraft II \cite{alphastarblog,vinyals2019grandmaster}. These systems have comparatively seen little use within the video game industry, and we believe lack of accessibility to be a major reason behind this. Indeed, really impressive results like those cited above are produced by large research groups with computational resources well beyond what is typically available within video game studios.

Our contributions are geared towards industry practitioners, by sharing experiments and practical advice for using RL with a different set of constraints than those met in the research community. For example, in the industry, experience collection is usually a lot slower, and there are time budget constraints over the runtime performance of RL agents. We thus favor off-policy algorithms to improve data efficiency by re-using past experience, and constrain our architectures to relatively small feedforward networks. The approach we propose in this paper is based on Soft Actor-Critic \cite{haarnoja2018soft}, which was originally designed for continuous actions problems. We explore ways to extend it to a hybrid setting with both continuous and discrete actions, a situation commonly encountered in video games. We also attempt to use normalizing flows \cite{normalizingflows2015} to improve the quality of the resulting policy with roughly the same number of parameters, and analyze why this approach may not be working as well as we initially expected. 

\section{Background and related work}

We consider the classical Markov Decision Process (MDP) setting where at each discrete time step $t$ the agent observes a state $s_t$ and must take an action $a_t \sim \pi(a_t | s_t)$, where $\pi$ is the agent's policy.
On the next time step, the environment transitions to the new state $s_{t+1} \sim P(s_{t+1} | s_t, a_t)$ and gives the agent a reward $r_t \sim P(r_t | s_t, a_t, s_{t+1})$.
The agent's objective is to find an optimal policy $\pi^*$ that maximizes the expected discounted return $\mathbb{E}_\pi\big[\sum_t \gamma^t r_t\big]$, where $\gamma \in [0, 1]$ is the discount factor.

In the following, we assume that a state is represented by a real-valued vector, in a format suitable to be provided as input to a neural network (e.g. with one-hot encoding of discrete state variables, and normalization of continuous features).
Actions may be either discrete, continuous, or a mix of both: a key contribution of this paper is to present a simple generic approach to action representation, suitable for most situations one may encounter when training game-playing agents.

\subsection{Soft Actor-Critic}

Soft Actor-Critic (SAC) \citep{haarnoja2018soft,haarnoja2018softapplications} is a state-of-the-art model-free algorithm that was originally proposed for continuous control tasks.
It is based on the idea of adding an entropy bonus to the objective optimized by the agent, i.e. maximizing $\mathbb{E}_\pi \Big[ \sum_t \gamma^t \big(r_t + \alpha \H(\pi(\cdot | s_t)) \big) \Big]$.
A higher $\alpha$ encourages the agent to take actions that are more random, which in particular can help with exploration. This $\alpha$ parameter can be learned during training by setting a target entropy for the policy \cite{haarnoja2018softapplications}.

% Explanation of the softmax, entropy framework, KL

\subsection{Normalizing flows}

Normalizing flows \cite{normalizingflows2015} are invertible transformations applied on top of an initial distribution to transform it into another distribution, usually with the goal of making it more expressive. The original SAC \cite{haarnoja2018soft} parameterizes the actor using a spherical Gaussian and uses the reparameterization trick to backpropagate through the parameters of the distribution. It is possible to apply normalizing flows on top of this Gaussian policy to make it more expressive \cite{mazoure2019leveraging}, while still being able to sample from the policy as well as compute the log-density at any point. This makes it possible to use normalizing flows to reparameterize the actor in SAC to get more complex policies while keeping the training algorithm unchanged.  

\cite{normflowsboostingtrpo} show that using an Inverse Autoregressive Flow (IAF) for on-policy trust region policy optimization can significantly improve exploration in high-dimensional tasks. \cite{ward2019improvingflows} use Real-valued Non Volume Preserving (Real NVP) flows to improve exploration in sparse reward settings, while \cite{haarnojaHSAC} use Real NVP flows to train maximum entropy policies in a hierarchical setting where each layer is trained on its own reward function. Our experiments with normalizing flows are similar to and inspired by \cite{mazoure2019leveraging}, who suggest that normalizing flows can be used to improve the expressiveness of policies in SAC to get a policy with the same level of quality using less parameters.  

As suggested in \cite{mazoure2019leveraging}, we use radial flows in our experiments. We sample $\varepsilon \sim \mathcal{N}(0, 1)$ (since we use the reparameterization trick), and denote by $h_{\theta}$ the function returning a sample from a Gaussian distribution with mean and standard deviation given by the policy parameterized by $\theta$. With $\{f_{\phi_i}\}_{i=1}^N$ the set of normalizing flows, we can sample from the policy as follows:

\begin{align*}
    w_0 &= h_{\theta}(\varepsilon, s_t) \\
    w_i &= f_{\phi_i} \circ f_{\phi_{i-1}} \circ ... \circ f_{\phi_{1}}(w_0) \\
    a_t &= \tanh(w_N)
\end{align*}

We denote by $q_0$ the density of the state-dependent Gaussian distribution $w_0$ is sampled from. The density of the policy is then tractable according to: 

\begin{equation}
    \log \pi(a_t, s_t) = \log q_0(w_0) - \sum_{i=1}^{N} \log \left| \det \frac{\partial f_{\phi_i}(w_{i-1})}{\partial w_{i-1}} \right|
\end{equation}

The equations corresponding to the radial flows are taken from \cite{normalizingflows2015} and can be found in the Appendix (Table \ref{nflows_parameterization}).

\subsection{Mixing discrete and continuous actions}

Most reinforcement learning research papers focus on environments where the agent's actions are either discrete or continuous.
However, when training an agent to play a video game, it is common to encounter situations where actions have both discrete and continuous components.
Typical examples include:
\begin{itemize}
    \item Playing with the same inputs as a player, whose controller may be equipped with both an analog stick (providing a range of continuous values) and buttons that can be pressed (yielding potentially many discrete actions through the various button combinations).
    \item Letting the agent choose among a set of high-level discrete actions (ex: move, jump, fire), each of them being associated with continuous parameters (ex: target coordinates for the move action, direction for the jump action, aiming angle for the fire action).
    \item Wanting the agent to control systems that have both discrete and continuous components, like driving a car by combining steering and acceleration (both continuous) with usage of the hand brake (a discrete binary action).
\end{itemize}

Such situations require algorithms that are able to handle a combination of discrete and continuous actions.
In what follows, we propose a parameterization of the policy that can be easily implemented in SAC, yielding a powerful generic off-policy RL algorithm for training game-playing agents.

% \subsubsection{Policy parameterization}

In order to deal with a mix of discrete and continuous action components, a first approach would be to use a fully continuous actor and somehow find a way to convert part of its continuous output into discrete actions \citep{vanhasselt2009,hausknecht2016,cianflone2019}.
Alternatively, one may use instead a fully discrete actor by discretizing the continuous actions, taking special care to prevent their number from exploding \citep{metz2017,andriotis2018,tang2019}.

% TODO cite https://medium.com/@kengz/soft-actor-critic-for-continuous-and-discrete-actions-eeff6f651954 ?

What we would like instead is a method that would naturally incorporate both discrete and continuous actions within the same algorithm (SAC) in a principled way.
In order to accommodate for the wide range of potential ways for an agent to interact with a video game environment, we generalize several existing ideas regarding action representation.
We first describe below our proposed generic setting, then relate it to specific examples from the literature.

We denote an agent action $a$ by a combination of discrete components $a^d = (a^d_1, \ldots, a^d_D)$ and continuous components $a^c = (a^c_1, \ldots, a^c_C)$. Each $a^d_i$ is an integer between $1$ and $K_i$, and represents the $i$-th discrete action that can be taken by the agent. Each $a^c_j$ is an $m_j$-dimensional continuous vector in $\mathcal{X}_j \subset \R^{m_j}$, and represents its $j$-th continuous action.
Discrete components are assumed to be independent given the observed state $s$, while continuous components are independent given both $s$ and the discrete actions, yielding the following decomposition:
\begin{equation*}
\begin{split}
\pi(a | s) & = \pi(a^d | s) \pi(a^c | s, a^d) \\
         & = \pi(a^d_1 | s) \ldots \pi(a^d_D | s) \pi(a^c_1 | s, a^d) \ldots \pi(a^c_C | s, a^d) \\
         & = \Pi_i \pi(a^d_i | s) \Pi_j \pi(a^c_j | s, a^d)
\end{split}
\end{equation*}
Here we slightly abuse notations by using the same letter $\pi$ to denote both discrete probability mass functions and probability density functions applied to different components of the action.
A rigorous treatment would rely on measure theory
%as e.g. in \citet{klimek2017}
but is beyond the scope of this paper.
We observe that many classical action representations fit the above decomposition:

\begin{enumerate}
    \item \label{singlediscrete} A single discrete action taken in the set ${1, \ldots, K}$, as in Atari games \citep{bellemare2013}.
    Here $D = 1$, $K_1 = K$ and $C = 0$.
    This yields $$\pi(k | s) = \pi(a^d_1 = k | s)$$
    
    \item \label{independent1D} $C$ independently sampled 1D continuous actions, as is typically done in continuous control tasks when computing $a^c_j = \tanh(\mu_j(s) + \varepsilon \sigma_j(s))$ with $\varepsilon$ sampled from a standard normal distribution \citep{haarnoja2018soft}.
    Here $D = 0$, $C$ is the total dimension of the continuous action space, and $m_j = 1$ for all $j$.
    This yields $$\pi(x | s) = \Pi_{j=1}^C \pi(a^c_j = x_j | s)$$
    
    \item A single $m$-dimensional continuous action vector, getting rid of the independence constraint from the previous case \#\ref{independent1D}.
    This can be achieved for instance with normalizing flows, where the continuous distribution being learned does not need to be be axis aligned anymore \citep{mazoure2019leveraging}.
    Here $D = 0$, $C = 1$ and $m_1 = m$.
    This yields $$\pi(x | s) = \pi(a^c_1 = x | s)$$
    
    \item \label{wei}An $m$-dimensional continuous action whose value should depend on a discrete action taken in $1, \ldots, K$, as proposed for parameterized action spaces by \citet{wei2018}.
    Here (in the general case with no independence assumption on the individual continuous components), this means that $D = 1$, $K_1 = K$, $C = 1$ and $m_1 = m$.
    This yields $$\pi(k, x | s) = \pi(a^d_1 = k | s) \pi(a^c_1 = x | s, a^d_1 = k)$$
    
    \item \label{paramactions} An alternative action representation for parameterized action spaces, where the agent takes a discrete action in $1, \ldots, K$, and each discrete action $k$ is parameterized by a different continuous $m_k$-dimensional vector.
    This is similar to what has been used e.g. by \citet{bester2019}.
    Here $D = 1$, $K_1 = K$, $C = K$ and $m_k$ is the dimension of the parameter being used when the discrete action is $k$.
    This yields $$\pi(k, x_k | s) = \pi(a^d_1 = k | s) \pi(a^c_k = x_k | s)$$
    The difference compared to the previous formulation \#\ref{wei} is that instead of using a single continuous parameter whose value depends on the discrete action being taken, we create multiple independent continuous parameters (one for each discrete action).
    Since each continuous parameter $a^c_k$ is only used when the agent takes its associated discrete action $k$, its value does not need to depend on the discrete action chosen by the agent, which is why $\pi(a^c_k = x_k | s)$ does not need to be conditioned on $a^d_1$.
    
    \item A set of $D$ discrete components, with each component $a^d_i$ ($1 \leq i \leq D$) being a discrete action taken in $(1, \ldots, K_i$). Such a representation has been used in particular by \citet{tang2019} to tackle continuous control tasks by discretizing each continuous dimension $i$ into $K_i$ discrete bins.
    In this example $D$ is the number of original continuous dimensions, $K_i$ is the number of bins in the discretization of the $i$-th dimension, while there are no continuous actions anymore ($C = 0$).
    This yields $$\pi(k_1, \ldots, k_D | s) = \Pi_{i=1}^D \pi(a^d_i = k_i | s)$$
    As motivated by \citet{tang2019}, such a representation avoids the exponential explosion of discrete actions that would occur if one chose to use instead a single discrete component as in \#\ref{singlediscrete}.
    Note that a similar idea is used in the action branching architecture of \citet{tavakoli2018}.

\end{enumerate}

\section{SAC with mixed discrete-continuous actions}

\subsection{Choosing an appropriate policy parameterization}

The examples from the previous section are a subset of all possible ways one can represent the action distribution over a mix of discrete and continuous components, using our generic proposed decomposition.
From a practitioner point of view, there is no single best representation that will fit all use cases.
For instance, if the agent needs to press buttons on a controller, and there are four buttons which can be set on/off, one can either consider a single discrete component with $2^4 = 16$ actions, or four independent binary discrete components.
The latter approach has the benefit of reducing the number of parameters that need to be learned, thanks to the factored representation, and thus generally scales better as the number of discrete components increases.
On the other hand, the independence assumption can make it harder for the agent to learn coordinated button presses, so the factored approach may perform badly when interactions between the discrete components really matter.
In general, we give the following advice to obtain an appropriate representation for a given task, based on our own experience:
\begin{itemize}
    \item Identify which action components (both discrete and continuous) should be made dependent of each other.
    When in doubt, it is advised to start with a simpler parameterization based on independent components, and only investigate later the potential benefits of more complex parameterizations.
    Note that in an MDP there always exists an optimal deterministic policy, for which all action components are independent given the state.
    As a result, it could be tempting to assume that everything can always be made independent (in order to simplify the model), but in practice this may slow down learning, in particular because it can prevent coordinated exploration across components (think of the above example with button presses).
    
    \item When a continuous component depends on a discrete component, consider duplicating it (one for each discrete action) as long as the model size remains reasonable: this will allow you to consider them as independent, making it easier for the model to specialize the value of the component to each discrete action.
    For instance, consider an $(x, y)$ continuous component which gives the 2D coordinates of a mouse click, where the agent has to select among several discrete actions before clicking (ex: attack, heal, follow): this continuous component may be replaced with three independent ones $(x_a, y_a)$, $(x_h, y_h)$ and $(x_f, y_f)$ associated to each discrete action, as in point \#\ref{paramactions} above.
    If this is too costly (due to a large number of discrete actions), you can instead (as in \#\ref{wei}) build the policy network in such a way that the continuous component head takes as input the discrete one: for details refer to \citet{wei2018}.
    
    \item If possible, try to avoid dependencies among continuous dimensions, so as to keep a simple parameterization where each action dimension can be sampled independently.
    For instance, if your continuous action is a pair $(a_x, a_y)$ giving the acceleration of your agent along the $x$ and $y$ axes, the agent may struggle to explore properly in situations where it needs to navigate narrow corridors that may not be axis aligned, since accelerations on both axes must be correlated to avoid bumping into the walls.
    In this specific case, one could for instance make the acceleration actions relative to the direction the agent is currently facing (by rotating the axes accordingly), making it easier for the agent to explore a wide range of forward accelerations without deviating from its trajectory.
    
\end{itemize}

%In the following, we provide guidelines on how to choose an appropriate representation, and how it translates in practice in the implementation of the SAC algorithm.

\begin{figure*}[htbp]
 
\begin{subfigure}{0.48\textwidth}
\includegraphics[width=0.9\linewidth]{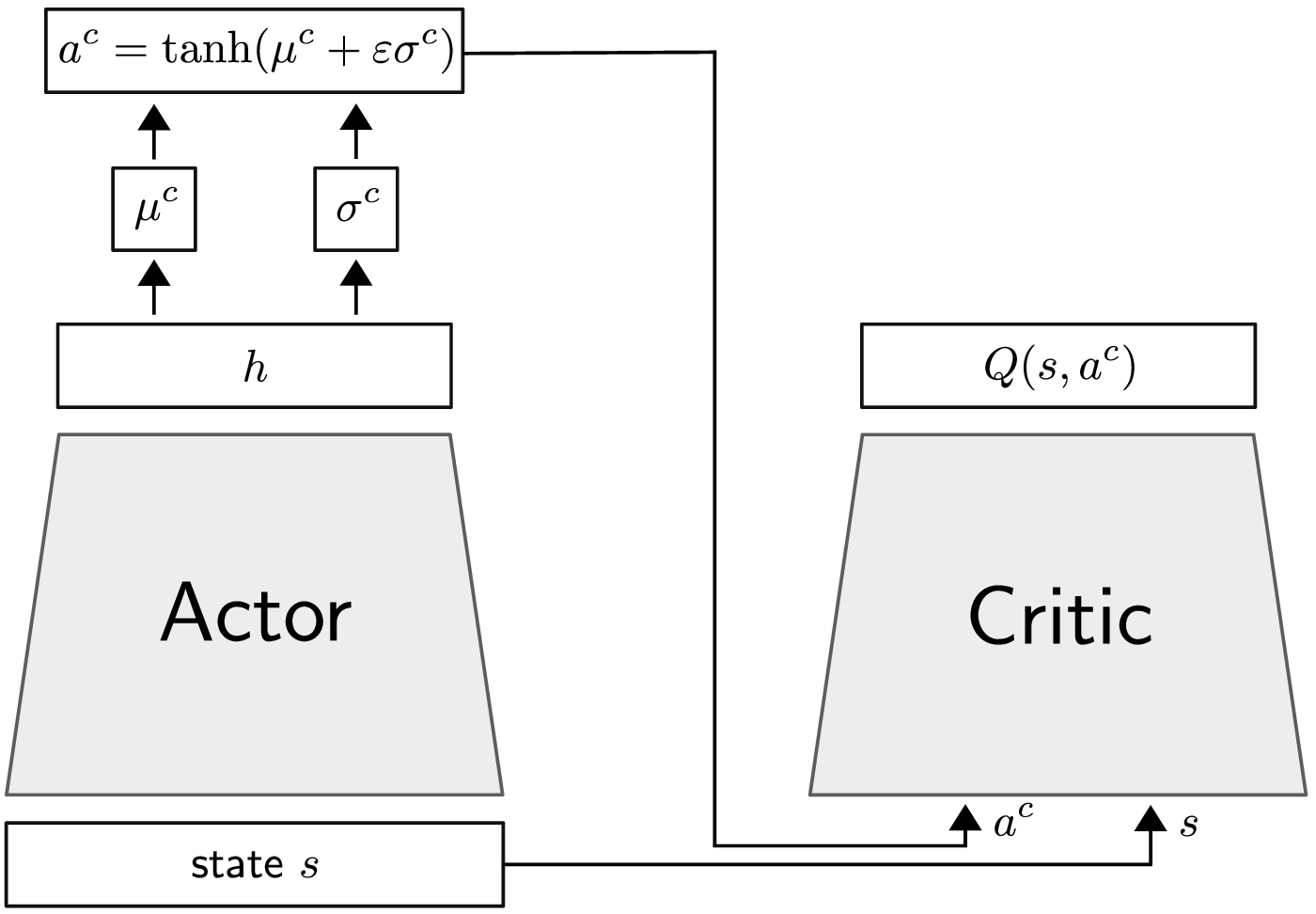}
\caption{Standard SAC}
\label{fig:sac_standard}
\end{subfigure}
\hfill
\begin{subfigure}{0.48\textwidth}
\includegraphics[width=0.9\linewidth]{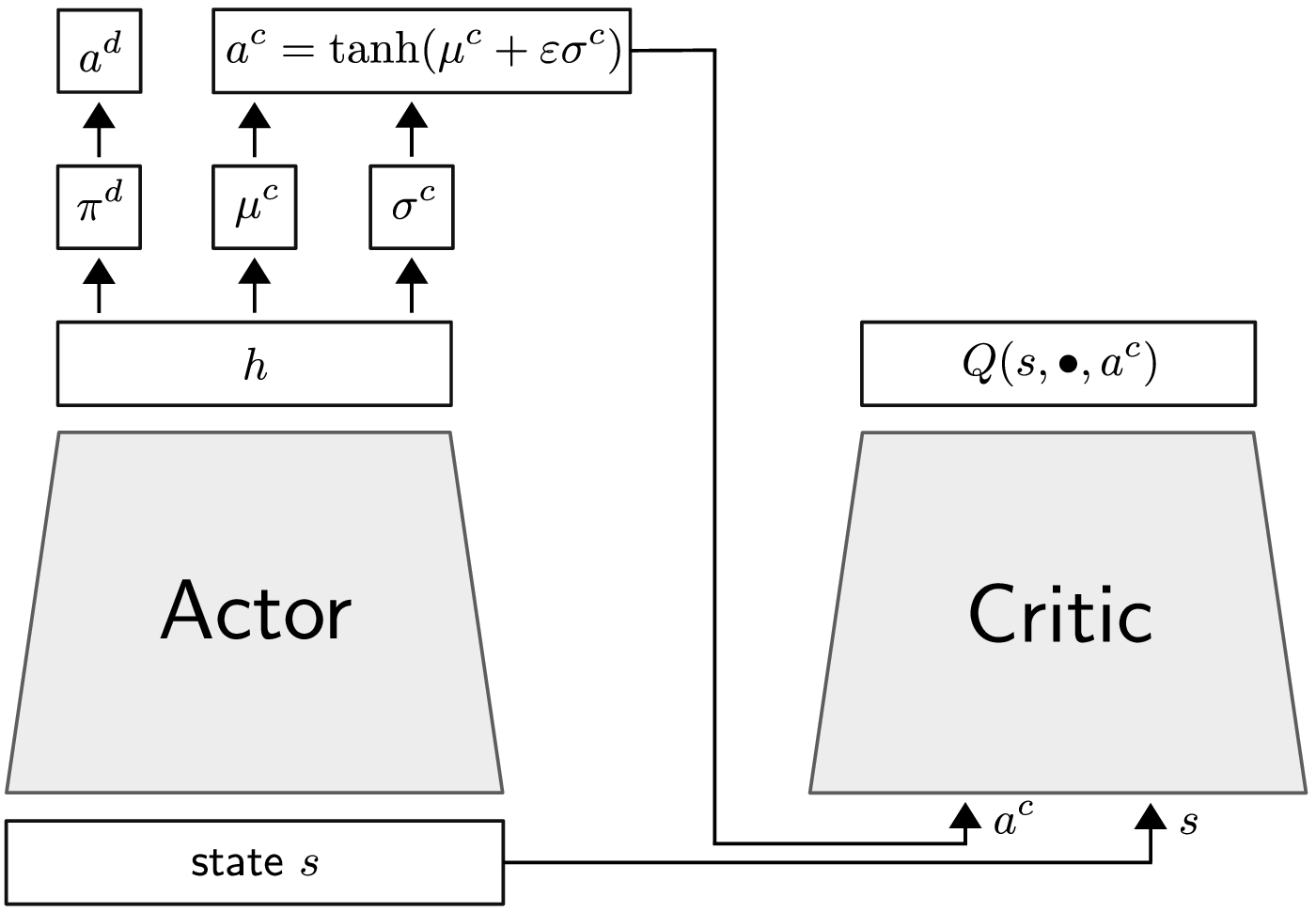}
\caption{Hybrid SAC (one discrete component and one continuous)}
\label{fig:sac_hybrid}
\end{subfigure}
 
\caption{(a) On the left, the standard SAC architecture for continuous actions. The actor outputs the mean and standard deviation vectors $\mu^c$ and $\sigma^c$ that are used to sample an action $a^c$ by injecting standard normal noise $\varepsilon$ and applying a $\tanh$ non-linearity (to keep the action within a bounded range).
The critic takes both the state $s$ and the actor's action $a^c$ to estimate their corresponding Q-value.
(b) On the right, an example of our proposed Hybrid SAC architecture, with two independent components (one discrete and one continuous).
The actor computes a shared hidden state representation $h$ that is used to produce both a discrete distribution $\pi^d$ (typically from a $\softmax$ layer) as well as the mean $\mu^c$ and standard deviation $\sigma^c$ of the continuous component.
The discrete action $a^d$ is sampled from $\pi^d$ while the continuous action $a^c$ is computed as in the standard SAC.
The critic network still takes both the state $s$ and the continuous action $a^c$ as input, but now predicts the Q-values of all discrete actions in its output layer.
}
\label{fig:sac}
\end{figure*}
 
\subsection{Practical implementation}
 
\subsubsection{Network architecture}
 
Fig.~\ref{fig:sac_standard} shows the typical architecture for the actor and critic networks used in standard continuous SAC implementations.
Using a different policy parameterization (like one of those described previously) calls for a different network architecture.
One common case is shown in Fig.~\ref{fig:sac_hybrid}, in the situation where the agent must take a combination of one discrete action $a^d$ with a set of independently sampled continuous parameters $a^c$.

Note that here, we chose to take an approach similar to \citet{xiong2018} where the critic's output layer contains the predicted Q-values of all discrete actions, instead of feeding the discrete action as input as done in the so-called ``multi-pass'' architecture of \citet{bester2019}.
This is because the former is the most commonly used architecture for discrete actions when using the popular Deep Q-Network algorithm and its variants \citep{mnih2015human,hessel2017}, but we acknowledge that the multi-pass architecture of \citet{bester2019} is also a valid alternative.
We actually implemented it in SAC, but our preliminary results did not show meaningful improvements, so we did not investigate it further at this time.

More complex policy parameterizations would lead to more elaborate architectures for the actor network than shown in Fig.~\ref{fig:sac_hybrid}, e.g.:
\begin{itemize}
\item In the case of multiple independent discrete components, the actor would output several corresponding discrete distributions $\pi^d_1, \ldots, \pi^d_D$.
\item If the continuous dimensions must be correlated, a different parameterization of $a^c$ may be used, for instance using normalizing flows \citep{mazoure2019leveraging}.
\item If the continuous action $a^c$ must depend on the discrete action chosen by the agent, then $a^d$ can be used as input when computing $\mu^c$ and $\sigma^c$ \citep{wei2018}.
\end{itemize}

\subsubsection{Learning algorithm}

The SAC algorithm \citep{haarnoja2018soft} is based on the idea of giving an entropy bonus proportional to the entropy of $\pi(a | s)$.
When the action has a discrete component,
the joint entropy definition yields $$\H\big(\pi(a^d, a^c | s)\big) = H\big(\pi(a^d | s)\big) + \sum_{a^d} \pi(a^d | s) \H\big(\pi(a^c | s, a^d)\big)$$
Although we could simply give a bonus proportional to this entropy, we argue that in some situations it may be beneficial to give different weights to its discrete and continuous parts.
This is because otherwise, depending in particular on the number of discrete and continuous actions, there would be a risk for one of these two entropies to ``overshadow'' the other, which could harm exploration.
As a result, we use as entropy bonus the weighted sum
\begin{equation}
\label{eq:entropy_bonus}
\alpha^d H\big(\pi(a^d | s)\big) + \alpha^c \sum_{a^d} \pi(a^d | s) \H\big(\pi(a^c | s, a^d)\big)
\end{equation}
where hyperparameters $\alpha^d$ and $\alpha^c$ encourage exploration for discrete and continuous actions respectively.
Note that these two hyperparameters can be tuned automatically during learning, using the same optimization technique as described in \citet{haarnoja2018softapplications}, by setting target values for the discrete and continuous parts in eq.~\ref{eq:entropy_bonus}.

%TODO explain the issue with parameters that can have a big entropy bonus when not used by all discrete actions

In terms of practical implementation, a list of the changes between our proposed Hybrid SAC and the original version can be found in the Appendix. Note that when there are only discrete actions, our approach is equivalent to the one proposed concurrently by \citet{christodoulou2019}.

\subsection{Experiments with parameterized actions}

We evaluate our Hybrid SAC implementation on the same three parameterized actions environments used by \citet{bester2019}:
\begin{itemize}
    \item {\em Platform} is a simple platformer-like game where the agent has three discrete actions (run, hop and leap), each associated with a 1D continuous parameter controlling the horizontal displacement.
    \item {\em Goal} is a soccer-based game where the agent needs to score a goal past a keeper that tries to intercept the ball. There are again three discrete actions, with respectively 2D, 1D and 1D continuous parameters.
    \item {\em Half Field Offense} is another soccer-based game, also with three discrete actions, but this time with respectively 2D, 1D and 2D continuous parameters.
\end{itemize}

In order to allow for a fair comparison with the state-of-the-art Multi-Pass Q-Network (MP-DQN) algorithm of \citet{bester2019}, we re-used their evaluation code and tried to match their hyperparameters whenever possible. We list the main remaining differences between our work and theirs in the appendix. 

Results are summarized in Table~\ref{tab:mixed}.
Both algorithms perform equally well on {\em Platform}, while MP-DQN exhibits slightly better performance on {\em Goal} and significantly better performance on {\em HFO}.
Note however that the MP-DQN results on {\em HFO} are based on an implementation that mixes Monte-Carlo returns with one-step returns to speed up convergence, an improvement that we did not implement in our Hybrid SAC.
The second row reports the performance of MP-DQN without mixing Monte-Carlo returns on {\em HFO}, showing that it degrades considerably (at least with the same hyper-parameters as MP-DQN).
%For this experiment, we re-used the same hyperparameters as those from \citep{bester2019}, so it might still be possible that better results could be obtained for MP-DQN (no MC) by further optimizing the algorithm's hyperparameters.

\begin{table}
\centering
\begin{tabular}{||c p{15mm} p{15mm} p{15mm}||} 
\hline
  & {\em Platform} & {\em Goal} & {\em HFO} \\ % [0.5ex] 
  & Return & P(Goal) & P(Goal) \\
\hline\hline
MP-DQN & $0.987 \newline \pm 0.039$ & $0.789 \newline \pm 0.070$ & $0.913 \newline \pm 0.070$ \\ 
\hline
MP-DQN (no MC) & - & - & $0.509 \newline \pm 0.110$ \\
\hline
Hybrid SAC & $0.981 \newline \pm 0.013$ & $0.728 \newline \pm 0.047$ & $0.639 \newline \pm 0.141$ \\
\hline
\end{tabular}
\caption{Comparison between the Multi-Pass Deep Q-Network (MP-DQN) algorithm from \citet{bester2019} and our Hybrid SAC implementation.
Mean performance with 95\% confidence interval is computed over 30 seeds.
Since the MP-DQN results on {\em HFO} take advantage of Monte-Carlo returns, while our Hybrid SAC does not, we also report in the second row the (significantly degraded) performance of MP-DQN without Monte-Carlo returns.
}
\label{tab:mixed}
\end{table}

While investigating potential reasons for the slightly worse average performance of Hybrid SAC on {\em Goal}, we realized that the entropy bonus from eq.~\ref{eq:entropy_bonus} may have an undesirable effect.
Discrete actions with a small $\pi(a^d | s)$ lead to a reduced entropy bonus for their associated $\pi(a^c | s, a^d)$.
This may cause the distribution of some continuous parameters to sometimes ``collapse''.
Our preliminary experiments with a variant aimed at avoiding this collapse matched the results of MP-DQN, but a more in-depth analysis of this variant is still needed before we can confidently report on its performance.

\subsection{Results in a commercial video game}

We trained a vehicle in a Ubisoft game, using the proposed Hybrid SAC with two continuous actions (acceleration and steering) and one binary discrete action (hand brake). The objective of the car is to follow a given path as fast as possible. A video of the resulting behavior is available at \url{https://youtu.be/bmrNMDEkPyQ}. Note that the agent operates in a test environment that it did not see during training, and that the discrete hand brake action plays a key role in staying on the road at such a high speed.
% WARNING: do not change the format of the link above, and do not include href or hyperref, it is specifically forbidden by AAAI.

\section{Experiments with Normalizing Flows}

Our main objective in an industry setting is to optimize the final performance of the policy under a budget constraint on its inference runtime. A potential avenue that could help is to augment the Gaussian policy obtained from a standard SAC algorithm with radial flows as advised in \cite{mazoure2019leveraging}, who report significantly improved performance with a reduced number of parameters. They suggest that such improvements could be related to the ability of the policy to be more expressive, for example by allowing it to be multimodal. In theory, training multimodal policies could yield agents that behave more naturally, for example in a driving situation where they could avoid an obstacle by turning either left or right.

\begin{figure*}[h]
\centering
\includegraphics[width=1\textwidth]{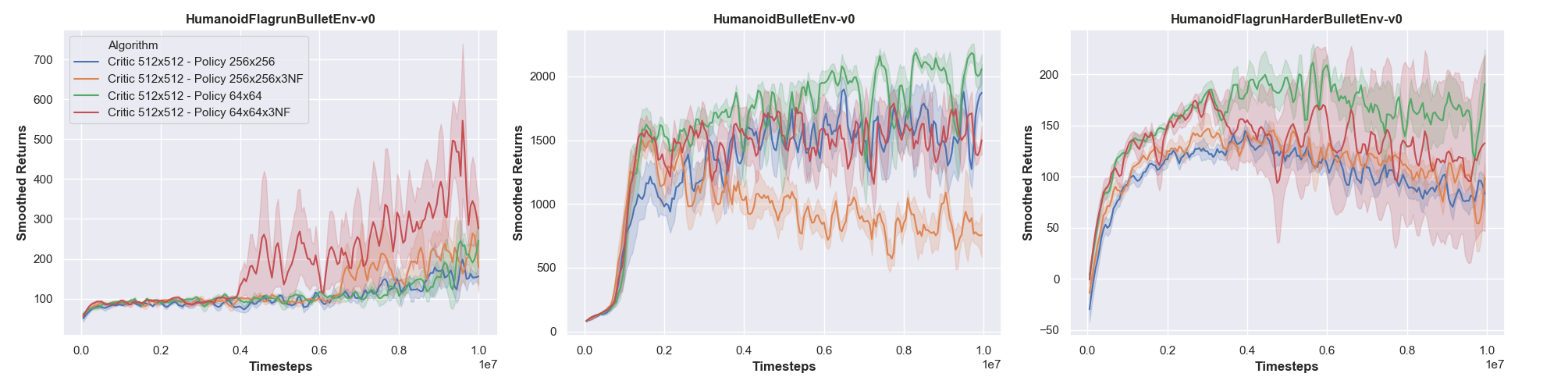} % Reduce the figure size so that it is slightly narrower than the column.
\caption{Comparison of the performance of SAC with and without radial normalizing flows on three Roboschool PyBullet environments. Curves are averaged on 5 random seeds, and smoothed using Savitzky-Golay filtering with window size 7.}
\label{fig_benchmark_roboschool}
\end{figure*}

\subsection{Bullet Roboschool benchmarks}

Our SAC baseline consists of a two-layer feedforward network outputting the mean and the standard deviation parameterizing a spherical Gaussian. The SAC-NF agent has the same architecture, but adds several radial flows on the output of the Gaussian. The resulting action is then squashed using a $\tanh$ as in \cite{haarnoja2018soft}. All the networks are trained using the Adam optimizer \cite{kingma2014adam}, details of the models' architectures can be found in appendix in Table \ref{robochool_hyperparameters}.

We evaluate the different architectures on the PyBullet Roboschool benchmark \cite{coumans2016pybullet}. We take one step of training every ten environment steps, and evaluate the policy every 50,000 steps. All the results are averaged over 5 random seeds. Since our intention is to see if the boost in policy expressiveness provided by normalizing flows can really help during training, we use bigger networks for the two critics so that the training is not limited by their relatively low capacity. Results of this comparison can be found in Fig.~\ref{fig_benchmark_roboschool}. 

Fig.~\ref{fig_benchmark_roboschool} shows that while a smaller policy with two hidden layers of 64 neurons with normalizing flows can get results that are competitive with bigger networks during the first million iterations as also reported by \cite{mazoure2019leveraging}, this advantage does not always hold as training goes further. Our results suggest that using normalizing flows on top of SAC does not yield a significant advantage compared to simply using the Gaussian policy of the baseline. In the following section, we will conduct an experiment on a toy environment to try to understand why.

\subsection{Normalizing Flows and SAC}

\begin{figure*}[h]
\centering
\includegraphics[width=0.7\textwidth]{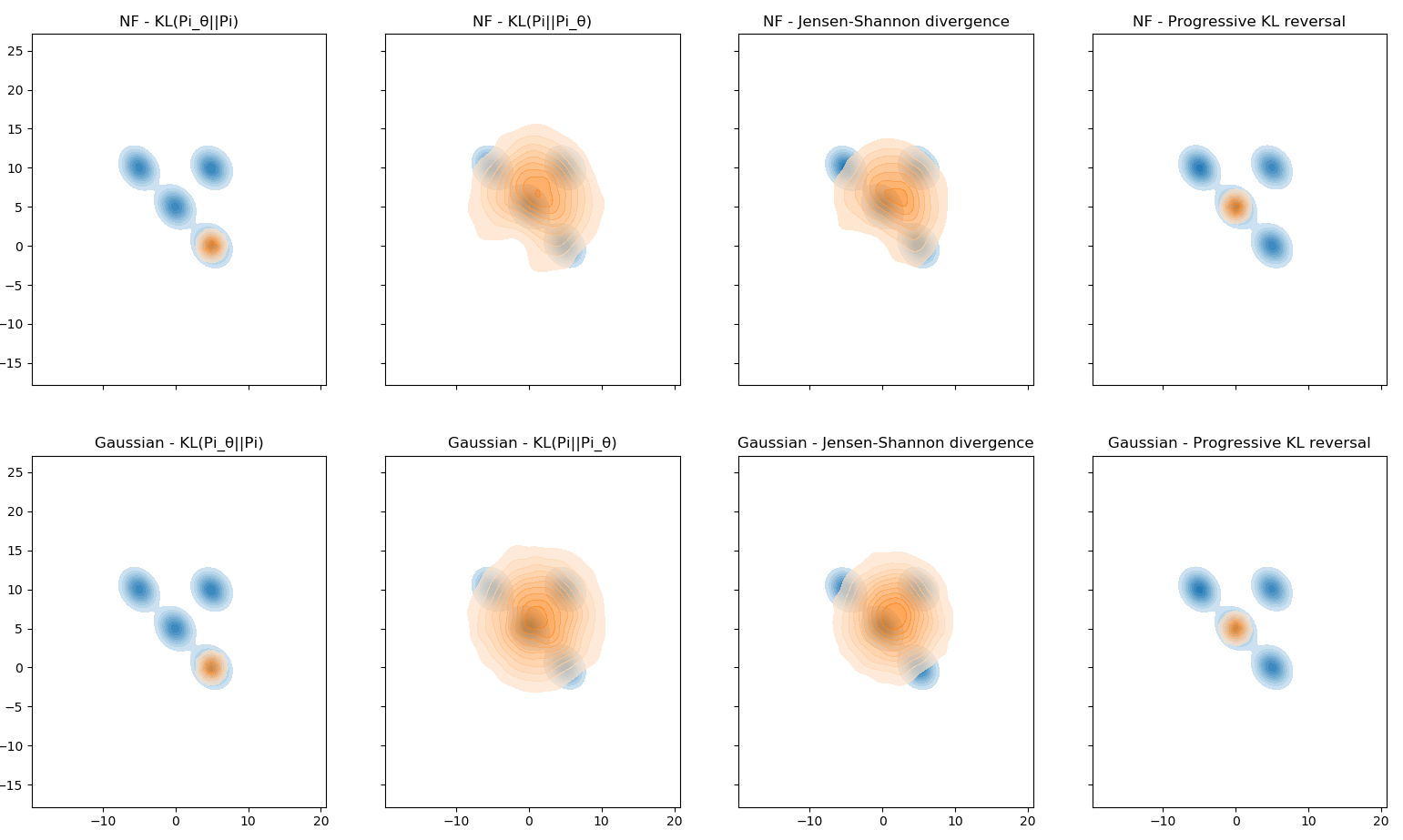} % Reduce the figure size so that it is slightly narrower than the column.
\caption{Comparison between the final shapes of the policy distribution with several objectives after trying to match a Gaussian mixture for 10,000 steps. The blue and orange densities correspond to the target Gaussian mixture $\pi$ and the learned distribution $\pi_{\phi}$ respectively.
Top row uses normalizing flows, while the bottom row is using a Gaussian policy.
Various divergence metrics are evaluated from left to right.
}
\label{fig_toy_distribution_matching}
\end{figure*}

In SAC, the actor tries to optimize the Kullback-Leibler (KL) divergence between the policy and a softmax on the soft Q-values with temperature $\alpha$. \cite{haarnoja2018soft} demonstrate that updating the policy in such a way improves it until convergence, and \cite{abdolmaleki2018relative} show that this update constrains the change of the policy. We thus try to minimize: 
\begin{equation} 
\label{SAC_KL_equation}
    J_{\pi}(\phi) = \mathbb{E}_{s_t \sim \pi_{\phi}} \left[ D_{KL} \left(\pi_{\phi}(\cdot | s_t) \bigg|\bigg| \frac{\exp\big(\frac{ Q_{\theta}(s_t, \cdot)}{\alpha}\big)}{Z_{\theta} (s_t)} \right) \right]
\end{equation}
where $Z_{\theta}$ is the partition function. However, the KL is not symmetric, and there is no theoretical ground in why $\pi_{\phi}$ should be the first argument. The main advantage of minimizing eq.~\ref{SAC_KL_equation} with $\pi_{\theta}$ in first position in the KL is tractability, as using the reparameterization trick allows us to minimize it without knowing the partition function. To do this, we rewrite the objective as an expectation on standard normal noise $\varepsilon$ and then sample this expectation: 
\begin{equation} 
\label{sampled_SAC_KL_equation}
    J_{\pi}(\phi) = \mathop{\mathbb{E}}_{\substack{s_t \sim \pi_{\phi} \\ \varepsilon_t \sim \mathcal{N}}} \Big[\log \pi_{\phi}\big(f_{\phi}(\varepsilon_t; s_t) | s_t \big) - Q_{\theta} \big( s_t, f_{\phi}(\varepsilon_t; s_t) \big) \Big]
\end{equation}
where $f_{\phi}$ reparameterizes the policy in terms of the noise $\varepsilon$. The particular choice of using the KL divergence from $\pi_{\phi}$ to the target softmax is motivated mainly by the convenience of its implementation. However, in classification tasks we generally try to minimize the negative log-likelihood, which is equivalent to minimizing the KL divergence from the empirical distribution to the parameterized one. In policy distillation, \cite{czarnecki2019distilling,parisotto2015actor,schmitt2018kickstarting} good results are reported when trying to minimize $\mathop{\mathbb{E}}_{\pi_{\phi}} \bigg [ \sum_{t=1}^{\tau} \nabla_{\phi}H^{\times}\big(\pi(s_t) || \pi_{\phi}(s_t) \big) \bigg ]$ where $H^{\times}$ is the cross-entropy, and the trajectories are sampled according to the student policy $\pi_{\phi}$ instead of the teacher policy $\pi$. We also did some experiments with distillation (not included here) which confirm that this way of doing policy distillation yields good results. All these observations suggest that if we interpret eq.~\ref{SAC_KL_equation} as trying to distill the ``teacher'' softmax over Q-values into the ``student'' parameterized policy $\pi_{\phi}$, a KL in the other direction would yield better results. This motivates the following comparison to measure the difference between these alternative objectives.

In this comparison, we fix a random state $s_0$ and try to get our policy to approximate a toy distribution $\pi$, in this case a Gaussian mixture, for this fixed state. Several objectives are evaluated, and we monitor the impact of using each objective on the shape of the final distribution after 10,000 steps of training. All hyperparameters are identical to our Roboschool experiment. We compare the Gaussian policy as used in SAC to the SAC-NF policy which adds three radial flows on top of the Gaussian. We compare the two directions of the KL divergence, as well as the Jensen-Shannon divergence \cite{lin1991divergence}. We also tried to linearly switch from $D_{KL}(\pi_{\phi} || \pi)$ to $D_{KL}(\pi || \pi_{\phi})$ during training. Results of this comparison can be found in Fig.~\ref{fig_toy_distribution_matching}. We use the kernel density estimate provided in the seaborn library \cite{waskom2018seaborn} to estimate the density of the distributions. 

From this toy experiment, one reason why normalizing flows did not seem to improve performance on Roboschool could be that any advantage gained in expressiveness of the policy by enriching it with normalizing flows is lost by the optimization procedure used in SAC. Indeed, when using the same objective as SAC (leftmost column in Fig.~\ref{fig_toy_distribution_matching}), there seems to be very little difference between using a Gaussian policy and one with normalizing flows, since both collapse on a single mode of the target distribution. Note that in this comparison we did not take the impact of the temperature into account (another comparison on the impact of the temperature can be found in Fig.~\ref{fig_alpha_comparison_sac_objective} and \ref{fig_alpha_comparison_reversed_sac_objective} in the Appendix). However, when we invert the KL (as we do in supervised learning and distillation) or use the Jensen-Shannon divergence, it appears that the normalizing flows help the policy better match the complete target distribution.

These results suggest that using normalizing flows could yield some benefits when used with other objectives than the one used in SAC. We ran some experiments reverting the KL using importance sampling, but training was too unstable. We identify the exploration of other metrics between distributions, such as the Jensen-Shannon divergence or the Wasserstein distance, as potential research avenues that could yield significant improvements when used in conjunction with normalizing flows in SAC.

\section{Conclusion}

We introduced Hybrid SAC, an extension to the SAC algorithm that can handle discrete, continuous and mixed discrete-continuous actions. It exhibits competitive performance with the state-of-the-art on parameterized actions benchmarks. We showed that Hybrid SAC can be successfully applied to train a car on a high-speed driving task in a commercial video game, demonstrating the practical usefulness of such an algorithm for the video game industry. Our study of the use of normalizing flows with the SAC algorithm also suggests that future approaches could further improve SAC by using other objectives than the KL, so as to better leverage normalizing flows.

\section{Acknowledgments}

We would like to thank the authors of \cite{mazoure2019leveraging} for insightful conversations and providing us with their implementation, as well as Paul Barde for his valuable feedback while writing this paper.   

\bibliography{bibliography}
\bibliographystyle{aaai}

\newpage

\section{Appendix}

%\subsection{Results in a real video game}

%We trained a vehicle in one of our games using the proposed Hybrid SAC, with two continuous actions (acceleration and steering) and one binary discrete action (hand brake). The objective of the car is to follow a given path as fast as possible. A video of the resulting behavior is available at \textit{https://youtu.be/bmrNMDEkPyQ} (Note that the agent operates in a test environment that it did not see during training).
% WARNING: do not change the format of the link above, and do not include href or hyperref, it is specifically forbidden by AAAI.

\subsection{Implementation differences between Hybrid SAC and SAC} 

\begin{itemize} \label{appendix_differences}
    \item The entropy bonus used in the target value for the critic network $Q$ is computed as in eq.~\ref{eq:entropy_bonus}, where the discrete part can be computed exactly (due to the finite number of discrete actions) while the continuous one needs to be approximated by sampling, as is usually done for continuous SAC.
    \item When optimizing the critic network $Q$ with a transition sampled from the replay buffer, only the output associated with the discrete action taken in this transition is optimized, similar to the Deep Q-Network algorithm \citep{mnih2015human}.
    \item The discrete part $\pi(a^d | s)$ of the policy is optimized by minimizing the KL divergence between this distribution and the one induced by the softmax on the Q-values with temperature $\alpha_d$.
    Since these Q-values depend on the continuous components $a^c$, we sample $a^c \sim \pi(a^c | s, a^d)$ in order to compute $q_d = Q(s, a^d, a^c)$ for each $a^d$, and take a gradient step to minimize the KL divergence between $\pi(a^d | s)$ and $P(a^d) \propto \exp(q_d / \alpha_d)$.
    As is usually done in continuous SAC, we multiply this gradient by $\alpha_d$ so as to prevent it from blowing up for small values of $\alpha_d$.
    \item Finally, the same $a^c$ sampled above are re-used to compute the update step for the continuous part of the policy. This update is essentially the same as in continuous SAC, as in eq.~7 of \citet{haarnoja2018softapplications}, except that it is performed as a weighted average over all discrete actions $a^d$, where the weight is given by $\pi(a^d | s)$ (i.e. mimicking the weighting scheme of eq.~\ref{eq:entropy_bonus}).
\end{itemize}

\subsection{Differences between MP-DQN and Hybrid SAC}

The main differences between our implementation of Hybrid SAC and the Multi-Pass DQN algorithm from \citet{bester2019} are the following:

\begin{itemize}
    \item We do not use the Multi-Pass architecture in our critic $Q$, because it significantly slows down learning and did not seem to really help in our preliminary experiments with SAC.
    Additional experiments are needed to fully investigate the potential benefits of this architecture for Hybrid SAC.
    \item For the sake of simplicity, we use a squashing $\tanh$ to bound the actions instead of the inverting gradient technique \citep{hausknecht2016}, and do not use gradient clipping.
    \item Since our approach is based on SAC, while MP-DQN is based on a combination of DQN \citep{mnih2015human} and DDPG \citep{lillicrap2016}, we do not use $\varepsilon$-greedy exploration nor add noise to continuous actions, but instead rely on the actor's stochasticity for exploration.
    \item We tweaked our actor and critic learning rates (as well as SAC-specific hyperparameters like the target discrete and continuous entropies) by a cursory search over reasonable-looking values.
    \item In the {\em Platform} environment, we do not use the custom initialization of the continuous parameters from \citet{bester2019} because we found it easy enough to get good results without it.
    \item In the {\em Half Field Offense} environment, we do not mix Monte-Carlo returns with one-step returns. Incorporating Monte-Carlo returns is not entirely straightforward in SAC due to the need to account for the entropy bonus, so we leave it to future work.
\end{itemize}

\begin{figure*}[!htbp]
\centering
\includegraphics[width=0.9\textwidth]{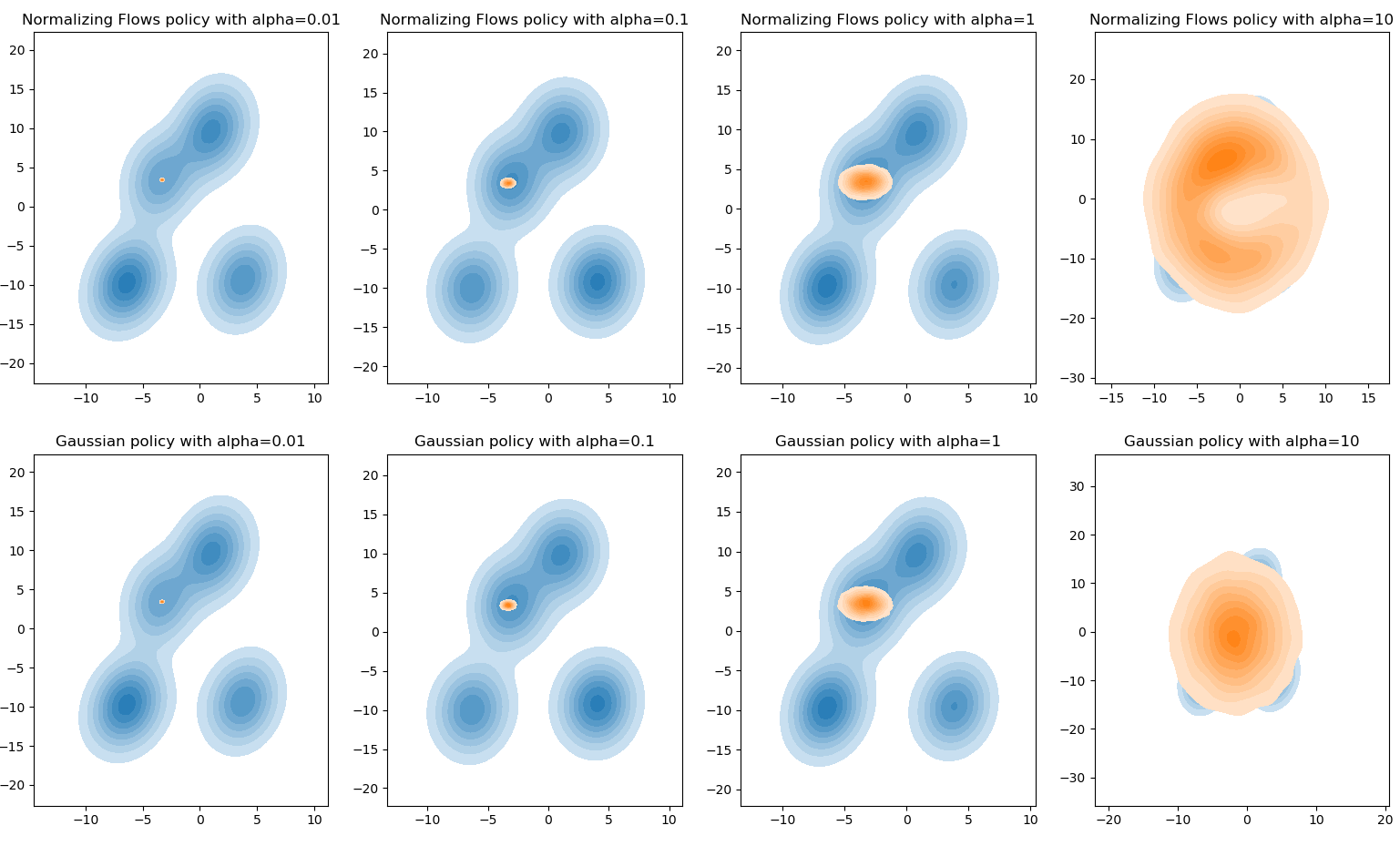} % Reduce the figure size so that it is slightly narrower than the column.
\caption{Comparison between the final shapes of the policy distribution $\pi_\phi$ (in orange) trained with the same objective as SAC after trying to match a Gaussian mixture $\pi$ (in blue) with different temperatures $\alpha$ for 10,000 steps. Note the collapse of the policies on one mode of $\pi$ unless $\alpha$ gets very high, for both normalizing flows (top row) and Gaussian policy (bottom row).
}
\label{fig_alpha_comparison_sac_objective}
\end{figure*}

\begin{figure*}[!htbp]
\centering
\includegraphics[width=0.9\textwidth]{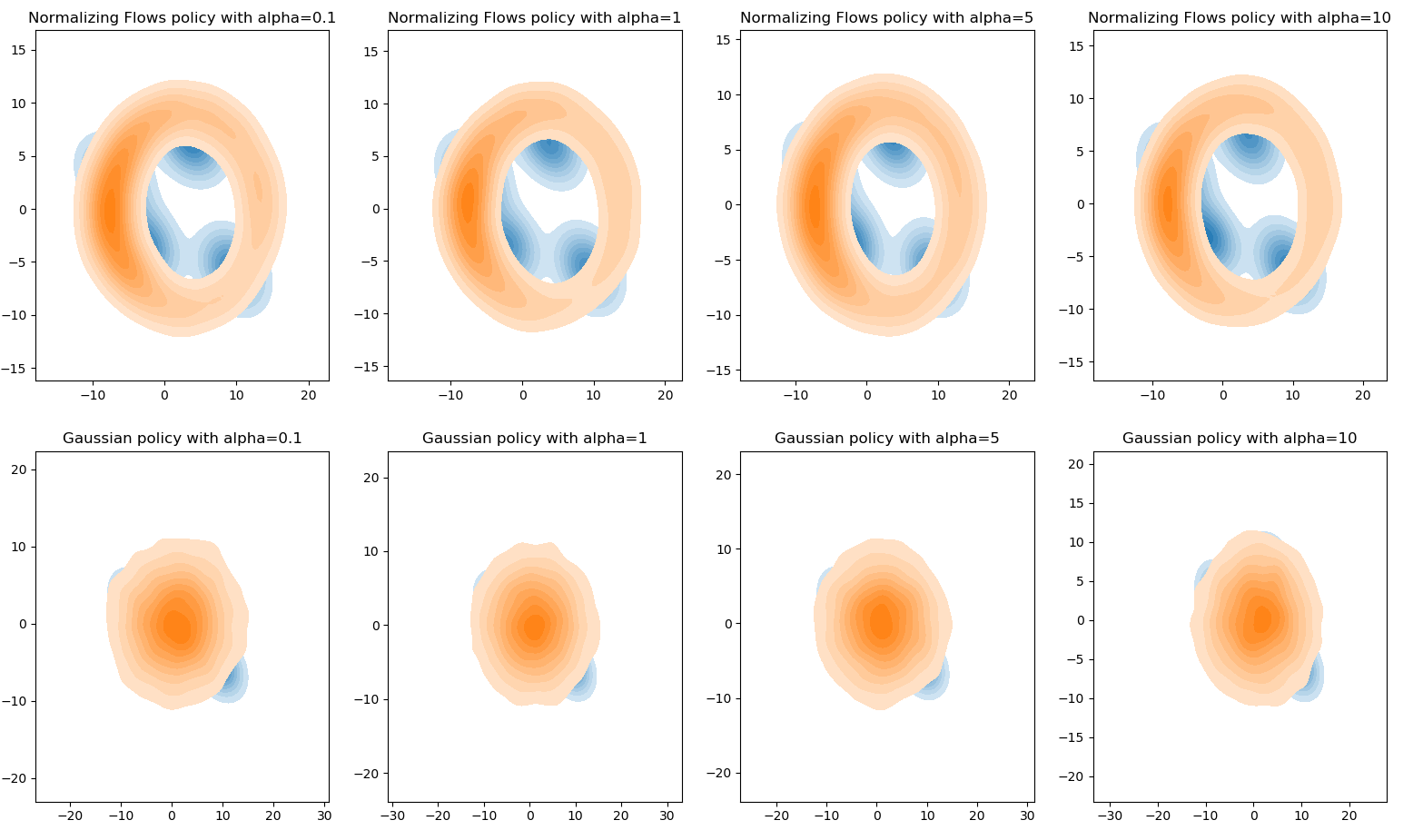} % Reduce the figure size so that it is slightly narrower than the column.
\caption{Same as Fig.~\ref{fig_alpha_comparison_sac_objective} but swapping the arguments of the KL divergence objective. Note that the policies no longer collapse onto a single mode of the target $\pi$, and the normalizing flow policy is better able to approximate the shape of $\pi$.}
\label{fig_alpha_comparison_reversed_sac_objective}
\end{figure*}

\subsection{Roboschool hyperparameters}

\begin{table}[!htbp]
\centering
 \begin{tabular}{c | c} 
 \hline
  Parameter & Value \\ % [0.5ex] 
  \hline
  Optimizer & Adam \\
  Learning rate & $3 \times 10^{-4}$ \\
  Discount ($\gamma$) & $0.99$ \\
  Replay buffer size & $10^6$ \\
  Alpha & $0.05$ \\
  Number of hidden layers & $2$ \\
  Neurons per hidden layer & $256$ \\
  Activation function & ReLU \\
  Minibatch size & $1024$ \\
  Target smoothing coefficient & $0.005$ \\
  Training / environment steps& $0.1$ \\
  Number of environment steps & $10^7$ \\
  Number of radial flows & $3$ \\
  \hline
\end{tabular}
\caption{SAC hyperparameters used in Roboschool.}
\label{robochool_hyperparameters}
\end{table}

\subsection{Normalizing flows parameterization}

We note $d$ the dimension of the action space. We parameterize $\phi=(z_0, x, y) \in \R^3$ as follows:

\begin{table}[!htbp]
\centering
 \begin{tabular}{c | c} 
 \hline
 Parameter & Value\\
 \hline
  $\alpha$ & $\exp(x)$ \\ 
  \hline
  $\beta$ & $-\alpha + \exp(y)$ \\ 
  \hline
  $f_{\phi}(z)$ & $z + \beta \cdot \frac{z - z_0}{\alpha + r(z)}$ \\ 
  \hline
  $r(z)$ & $||z - z_0||_2$ \\
  \hline
  $\det \frac{\partial f_{\phi}(z)}{\partial z}$ & $\left( 1 + \frac{\beta}{\alpha + r(z)} - \frac{\beta r(z)}{(\alpha + r(z))^2} \right) \cdot \left( 1 + \frac{\beta}{\alpha + r(z)} \right)^{d - 1}$ \\ 
  \hline
\end{tabular}
\caption{Parameterization of the normalizing flows.}
\label{nflows_parameterization}
\end{table}

\subsection{Additional experiments with normalizing flows}

Fig.~\ref{fig_alpha_comparison_sac_objective} and \ref{fig_alpha_comparison_reversed_sac_objective} extend the results presented in Fig.~\ref{fig_toy_distribution_matching} where we train a policy $\pi_\phi$ to match a target distribution $\pi$.

\end{document}